%% file: ijcai24.tex
\documentclass{article}
\usepackage{ijcai24}

\usepackage{times}
\usepackage{soul}
\usepackage{url}
\usepackage[hidelinks]{hyperref}
\usepackage[utf8]{inputenc}
\usepackage[small]{caption}
\usepackage{graphicx}
\usepackage{amsmath}
\usepackage{amsthm}
\usepackage{booktabs}
\usepackage{algorithm}
\usepackage{algorithmic}
\usepackage[switch]{lineno}
\usepackage{hyperref}
\usepackage{cleveref}
\usepackage{commath}
\usepackage[absolute,overlay]{textpos}
\usepackage{multirow}
\usepackage[pscoord]{eso-pic}

\creflabelformat{equation}{#2\textup{#1}#3}

\usepackage{tikz}
\def\checkmark{\tikz\fill[scale=0.4](0,.35) -- (.25,0) -- (1,.7) -- (.25,.15) -- cycle;}

\newcommand{\placetextbox}[3]{
  \setbox0=\hbox{#3}
  \AddToShipoutPictureFG*{
    \put(\LenToUnit{#1\paperwidth},\LenToUnit{#2\paperheight}){\vtop{{\null}\makebox[0pt][c]{#3}}}%
  }%
}%

\title{A Survey on Efficient Federated Learning Methods for Foundation Model Training}

\author{
    Herbert Woisetschl\"ager$^1$
    \and
    Alexander Erben$^1$\and
    Shiqiang Wang$^2$ \and 
    \\
    Ruben Mayer$^3$ \And 
    Hans-Arno Jacobsen$^4$
    \affiliations
    $^1$Technical University of Munich \\
    $^2$IBM Research \\
    $^3$University of Bayreuth \\
    $^4$University of Toronto
    \emails
    \{herbert.woisetschlaeger, alex.erben\}@tum.de,
    wangshiq@us.ibm.com,
    ruben.mayer@uni-bayreuth.de,
    jacobsen@eecg.toronto.edu
}

\newcommand{\mycite}[1]{\citeauthor{#1}~[\citeyear{#1}]}

\begin{document}
    \maketitle

    \begin{abstract}
        Federated Learning (FL) has become an established technique to facilitate privacy-preserving collaborative training across a multitude of clients. 
        However, new approaches to FL often discuss their contributions involving small deep-learning models only and focus on training full models on clients.
        In the wake of Foundation Models (FM), the reality is different for many deep learning applications.
        Typically, FMs have already been pre-trained across a wide variety of tasks and can be fine-tuned to specific downstream tasks over significantly smaller datasets than required for full model training.
        However, access to such datasets is often challenging.
        By its design, FL can help to open data silos.
        With this survey, we introduce a novel taxonomy focused on computational and communication efficiency, the vital elements to make use of FMs in FL systems.
        We discuss the benefits and drawbacks of parameter-efficient fine-tuning (PEFT) for FL applications, elaborate on the readiness of FL frameworks to work with FMs and provide future research opportunities on how to evaluate generative models in FL as well as the interplay of privacy and PEFT.
    \end{abstract}

    \placetextbox{0.5}{0.95}{
    \fbox{
        \small
        Accepted for publication at IJCAI 2024. Please find the paper in the official proceedings: \url{https://doi.org/10.24963/ijcai.2024/919}
        }
    }%


    \section{Introduction}
    \label{sec:intro}
    \input{chapters/01_introduction}

    \section{Taxonomy}
    \label{sec:taxonomy}
    \input{chapters/02_taxonomy}

    \section{Computational Efficiency}
    \label{sec:fl_lmm_fm}
    \input{chapters/03_learning_frameworks}

    \section{Communication Efficiency}
    \label{sec:fl_communication}
    \input{chapters/04_importance_of_communication}

    \section{Are FL Frameworks Ready for FMs?}
    \label{sec:fl_systems}
    \input{chapters/05_fl_systems_with_fm_and_llm}

    \section{Related Work}
    \label{sec:related-work}
    \input{chapters/06_related_work}

    \section{Conclusions \& Future Directions}
    \label{sec:conclusions}
    \input{chapters/07_conclusion}

    \newpage
    \appendix

    \section*{Acknowledgments}
    This work is partially supported by the Bavarian Ministry of Economic Affairs, Regional Development and Energy (Grant: DIK0446/01).

    {
        \small
        \bibliographystyle{named}
        \bibliography{ijcai24}
    }
    
\end{document}

%% file: chapters/01_introduction.tex
Foundation Models (FMs) \cite{bommasani2021} have conquered the deep learning world with unprecedented speed, enabling generative artificial intelligence for a broad audience. 
As FMs have been pre-trained on an extensive data basis and can be used in multi-modal applications, they perform well over a wide range of tasks.
To specialize these models on a downstream task, we use fine-tuning that can either be done over the full model or with parameter-efficient fine-tuning techniques (PEFT) \cite{Hu2021,Lester2021,zaken_2021_bitfit}.
A major advantage is that fine-tuning requires orders of magnitude smaller datasets than pre-training but benefits from access to a variety of samples pertaining to a task.

Access to a breadth of data has always been challenging in deep learning, as data owners are typically reluctant to share their data with service providers.
To tackle the data access challenge, \mycite{mcmahan2017} introduced \emph{Federated Learning} (FL). 
FL enables privacy-preserving machine learning over decentralized data without the necessity of sharing input data and does not require high bandwidth client connections.
Rather, a set of clients collectively train a model and send their local model updates to a server that subsequently aggregates the updates to a global model.
In FL applications that involve small models with less than 1 million parameters, we may spend as much time on communication as we spend on computation \cite{Yousefpour2023_green_fl}. 
However, it is desirable to design systems in such a way that computation takes up the majority of time.
Luckily, the larger the models become, the time spent on training is larger than on communication \cite{ryabinin2023,woisetschlaeger2023,beutel2020_flower}. 
As such, scaling the model size in FL systems can be desirable, and this introduces beneficial properties that can aid us in training large models with several 100 million parameters and beyond.
These properties render FL the perfect choice for fine-tuning FMs for downstream tasks.
FL can provide access to a significantly larger and more diverse data basis while benefiting from the increased time spent on the computation of FMs over small models. 
Yet, the costs of transmitting model updates remain significant even with the increased computational load of FMs \cite{Yousefpour2023_green_fl}, making it a priority to jointly optimize the training and communication efficiency. 

Our survey is the first to study advances in computational and communication-efficient methods for \emph{FM training} in FL applications.
Our work contains three distinct contributions: 

\begin{itemize}
    \item \textbf{A novel taxonomy on FL methods for FM training focused on the core challenges in computation and communication}. 
    We discover a gap between FL methods to increase computational efficiency and techniques to improve communication efficiency. While we see research on computational efficiency for FM training and fine-tuning in FL applications, communication efficiency methods are predominantly tailored to full-model training. 
    Our taxonomy aims to identify synergies between FL methods for FMs and efficient communication methods. 

    \item \textbf{Holistic evaluation of existing FL computational efficiency methods for FMs and communication efficiency techniques}.
    We study how existing techniques can help drive the adoption of FMs in FL applications and what needs to be done to render FL frameworks ready for large models. 

    \item \textbf{We thoroughly discuss future research directions}. 
    We highlight future directions for research on computational and communication efficiency as these domains grow closer together.
    Also, we show what is necessary to make FMs in FL applications a reality, especially with regard to generative tasks and privacy considerations.
    
\end{itemize}

\begin{figure}[!ht]
    \centering
    \includegraphics[height=4.8cm]{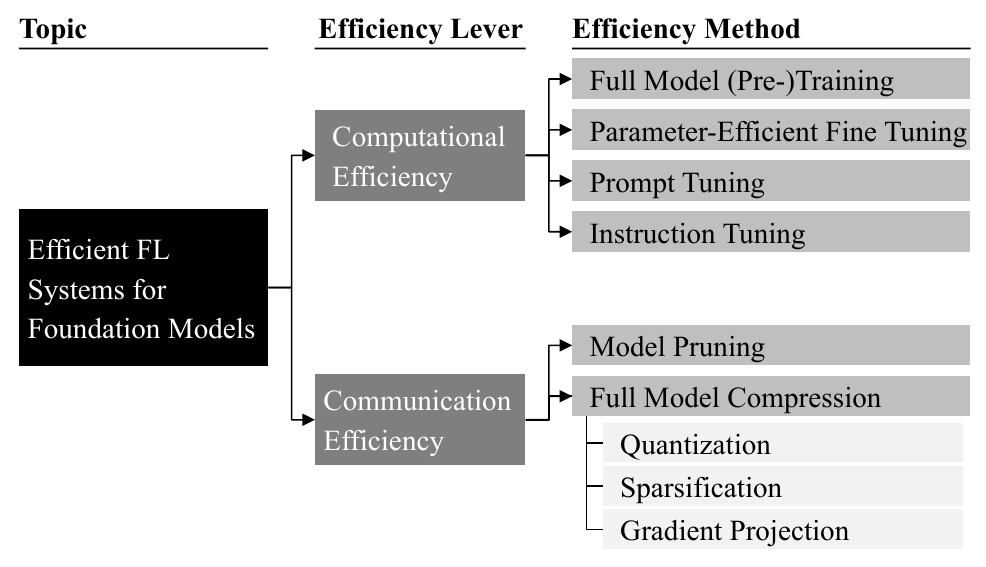}

    \caption{Our Taxonomy. Foundation Models, in conjunction with Federated Learning, require efficient computational and communication methods.}
    \label{fig:taxonomy}
    \vspace{-8pt}
\end{figure}

%% file: chapters/02_taxonomy.tex
Our taxonomy introduces a novel perspective focused on the current developments in the field of efficient computational and communication methods for FL applications intended for training and fine-tuning FMs. While communication efficiency has been studied extensively in FL, computational advancements in conjunction with large models are currently emerging. Our taxonomy is visualized in \Cref{fig:taxonomy}.

\subsection{Basics of Federated Learning}
To understand the relevancy of our taxonomy, it is key to briefly introduce the fundamentals of FL and the notations of this paper.

Usually coordinated by a server, the overall goal of FL is to collaboratively and iteratively train a DL model across a set of clients $N$ and minimize the global loss, where $f$ denotes an objective function with parameters $w$, 
\begin{equation}
    \min_w \, f(w) := \frac{1}{|N|} \sum_{n=1}^{|N|} f_{n}(w)\mathrm{.}
\end{equation}

With this, we create a model that generalizes across all clients $n \in N$.
Specifically, at the beginning, we commonly initialize the model weights across clients. Then, clients train the model on their local data and return the updated model parameters to the server.
At the same time, each client updates its local parameters with fixed minibatch size $m$ over one or more epochs by applying gradient descent steps $\nabla l(w^n_t;m)$ to the model, 
\begin{equation}
    w_{t+1}^{n} = w_t - \eta \nabla l(w^n_t;m), \, \forall n \in N\mathrm{.}
\end{equation}

Subsequently, only the model parameters $w_{t+1}^n$ are communicated back to the server.

\subsection{Taxonomy Explanation}

Our taxonomy is centered on two major challenges in FL: the computational and communication efficiency levers. 
While these challenges have been studied in FL extensively in the past \cite{Zhang2021}, existing approaches predominantly focus on small models with $< 10\mathrm{M}$ parameters. 
With the emergence of FMs as the backbone for multi-modal applications, we need to combine computational and communication efficiency in FL as these FMs typically come with $> 1~\mathrm{billion}$ parameters \cite{zhang2023_fedgpt}; a growth factor of $100\times$
This introduces additional computational load for FL clients, while their resources are often scarce already \cite{beutel2020_flower}.
At the same time, communication loads are also growing since many parameters need to be transmitted. 

\paragraph{Computational Efficiency.}
We discuss computational efficiency levers along four major categories. 
Full model training is used to train large transformer models from the very beginning (\Cref{sec:full_model_training}). 
Parameter-efficient fine-tuning techniques can be utilized to improve a pre-trained FM for a specific downstream task (\Cref{sec:peft}). Prompt tuning enables performance improvements of an FM without training the model itself but by designing textual prompts that we prepend to an input (\Cref{sec:prompt_tuning}). 
Instruction tuning enables fine-grained control over the model training process and allows for a high degree of model specialization for certain downstream tasks (\Cref{sec:instruction_tuning}).

\paragraph{Communication Efficiency.} While computationally efficient methods for FM training in FL applications can already reduce the number of parameters to communicate, there is still a significant amount of data to be communicated between clients and servers. For instance, parameter-efficient fine-tuning methods only require 1--2\% of parameters to be trainable. This still amounts to up to 14M parameters when working with Alpaca-7B \cite{zhang2023_fedgpt}, larger than the majority of models (with $< 1\mathrm{M}$ trainable parameters) currently being discussed in FL research \cite{He2020_fedml,beutel2020_flower}. 

We discuss communication efficiency methods along two major categories. (I) \emph{Model pruning} is a method to communicate parts of a model between clients and the server, which resemble the most important parameters for a client (\Cref{sec:pruning}). (II) \emph{Full model compression} is divided into three sub-categories: First, quantization is a method to decrease the numeric precision of model parameters. Second, sparsification is used as a way to zero out less important model parameters. Third, gradient projection transforms high dimensional parameter matrices in client updates into scalar vectors for communication (\Cref{sec:full_model_compression}).

\begin{table*}[!ht]
    \centering
    \resizebox{0.85\textwidth}{!}{
        \input{tables/Fine_Tuning_FL_Table}
    }
    \caption{
    Computational Efficiency Methods for FL Systems and FM. 
    FMs are generally multi-modal and provide a strong performance across a variety of domains that are well explored in centralized learning but not in FL \protect\cite{Dosovitskiy2020,gpt4_paper,yan2023_uniaudio}. 
    }
    \label{tab:summary_computation_works}
    \vspace{-3pt}
\end{table*}

%% file: tables/Fine_Tuning_FL_Table.tex
\begin{tabular}{l|llr|rr|ccc}
    \toprule
    \multicolumn{1}{l}{}    & \multicolumn{1}{c}{Training}   & \multicolumn{1}{c}{Underlying FL}                & \multicolumn{1}{c}{Max. Model}    & \multicolumn{1}{c}{Trainable} & \multicolumn{1}{c}{Comms Savings}     & \multicolumn{3}{c}{Domain} \\
    \multicolumn{1}{l}{Paper}   & \multicolumn{1}{c}{Regime}     & \multicolumn{1}{c}{Aggregation Strategy} & \multicolumn{1}{c}{Params}        & \multicolumn{1}{c}{Parameters}    & \multicolumn{1}{c}{compared to FMT}      & CV & NLP & Audio \\
    \midrule
    \textit{Centralized Learning}   & FMT, PEFT, PT, IT         & --                                        & $\geq 7\mathrm{B}$                 & --                            & --                              & \checkmark    & \checkmark    & \checkmark \\
    \midrule
    FedBERT \cite{Tian2022}        & FMT                       & Weighted Average                           & 117M                              & 110M                          & 0\%                              &               & \checkmark    &              \\
    FedCLIP \cite{Lu2023_fedclip}  & PEFT                      & Weighted Average                           & 85M                               & 530K                          & $>99\%$                              & \checkmark    &               &          \\
    FedPEFT \cite{sun2023}         & PEFT                      & Weighted Average                           & 85M                               & 230K                          & $>99\%$                              & \checkmark    &               &          \\
    FedPETuning \cite{Zhang2023}   & PEFT                      & Weighted Average                           & 125M                              & 1.25M                         & $99\%$                              &               & \checkmark    &           \\
    SLoRA \cite{Babakniya2023}     & PEFT                      & Weighted Average                           & 67M                               & 140K                          & $>99\%$                              &               & \checkmark    &          \\
    FedDPA \cite{Yang2024DualPersonalizingAF} & PEFT & Weighted Average & 7B & N/A & $> 99\%$ & & \checkmark & \\
    FedPrompt \cite{zhao2022}      & PT                        & Weighted Average                           & 223M                              & --                            & $>99\%$                              &               & \checkmark    &          \\
    FedIT \cite{zhang2023_fedgpt}  & IT                        & Weighted Average                           & 7B                                & --                            & $>99\%$                              &               & \checkmark    &          \\       
    \bottomrule
    \bottomrule

\end{tabular}

%% file: chapters/03_learning_frameworks.tex
This section discusses recent methods to train FMs with FL methods. 
We distinguish between full model training (FMT), as it has been studied frequently in the FL domain, PEFT, prompt tuning (PT), and instruction tuning (IT). \Cref{tab:summary_computation_works} summarizes existing computational efficiency methods for FM training.

\subsection{Full Model Training}
\label{sec:full_model_training}
Generally, full model training is referred to when training all parameters of a neural network. In this approach, we locally train a model on all clients with the objective of minimizing the loss $l$.

The BERT model is one of the first models to use the transformer layer architecture  --~the building blocks of FMs~-- to achieve state-of-the-art performance at the time of its release. \mycite{Tian2022} discuss federated pre-training of BERT-family models with up to 117M parameters by applying masked language modeling (MLM). In MLM, the loss is defined over the sum of probabilities $P$ of predicted tokens $\hat{x}$ over the representation function of a masked sentence $g\big(\frac{x}{M(x)}\big)$, 
\vspace{-8pt}
\begin{equation}
    \label{eq:mlm_loss}
    l(w^t_n,m) = - \sum_{\hat{x} \in \mathcal{M(\mathrm{x})}} \mathrm{log} P\left(\hat{x}\left|g\left(\frac{x}{M(x)}\right)\right.\right), \, \forall n \in N \mathrm{.}
\end{equation}

In FL applications, supervised learning can be challenging as we cannot ensure a proper data labeling process for supervised learning because we usually cannot access client data. 
Here, MLM can be beneficial since it is a self-supervised learning technique that masks parts of a sentence. Those masked parts will be used as the prediction target to automatically create an input and target sample.
Federated pre-training provides a net benefit over training on local datasets \cite{Ding2023_nature,Babakniya2023,He2020_fedml}. However, the perplexity, an indicator for quality in large models, yields four orders of magnitude worse results for large models (e.g., GPT-2) trained in a federated fashion than centralized training, which is a stark indicator of low model quality \cite{Tian2022,radford2019language}. Nonetheless, for use cases involving sensitive data and strict privacy regulation, full model pre-training allows the creation of foundation models based on federated data.

While FL pre-training has shown some promise, it is brittle and has shown to be worse at model sizes above $100\mathrm{M}$ parameters \cite{Tian2022}. the applicability of model pre-training is currently limited due to the significantly lower model quality than in centralized training.

\subsection{Parameter Efficient Fine-Tuning}
\label{sec:peft}
Generally, PEFT is used to improve the performance of large models already trained on a large data corpus further and provide good performance across various tasks.
This is especially effective since the data required for fine-tuning is orders of magnitude smaller than for pre-training. When applying PEFT, additional fully connected layers are inserted into the pre-trained model between the transformer blocks.  While the original model weights are frozen,  only the newly added layers are trained, typically resulting in $\geq 98\%$ less communication \cite{Hu2021}.
This renders PEFT techniques well-suited for FL applications since they address computation and communication alike.
However, \mycite{Babakniya2023} show in their study that PEFT is more sensitive towards non-IID data than FMT, but this sensitivity can be mitigated.
PEFT can be applied in FL applications as follows. 

\paragraph{Sparse fine-tuning of pre-trained model parameters.} As communication is a key concern in FL, reducing the number of parameters to communicate between client and server has become a priority. 
One of the most used approaches to achieve this is BitFit \cite{zaken_2021_bitfit}. 
The technique freezes almost all model parameters $w_t$ and only trains the bias term $b$ and the final classification layer $w^{\mathrm{final}}_{t}$ over the input features $a_t$, where the next-layer input features $a_{t+1} = a_{t} \cdot w_{t} + b$.
With this technique, it is only required to communicate $b$ and $w^{\mathrm{final}}_{t}$. The communication load is reduced by $\geq 99\%$, i.e., instead of communicating 100 million parameters, only 100 thousand parameters are sent over the network.

\mycite{sun2023} propose FedPEFT, a framework for federated transformer fine-tuning that freezes the model weights to retain upstream knowledge within the model and adjust the systematic error for the downstream task. Their experimental results on vision transformers (ViT-B with 85M parameters) show on-par performance compared to full model fine-tuning on non-IID data, all the while reducing communication by~99.8\%.

\paragraph{Adapter-based fine-tuning of additionally added parameters.} When aiming to maintain a pre-trained model while introducing task-specific knowledge, adapter-based fine-tuning techniques provide strong performance and on-par efficiency compared to sparse fine-tuning techniques \cite{Houlsby2019}. Here are two adapter layers introduced in each transformer block of a foundation model. The adapter layer $a^{a}_{t+1}$ is calculated based on a downstream projection $w_{t}^{\mathrm{down}} $ of input feature $a^{a}_t$ into a lower-dimensional space $w_t^{\mathrm{down}}  \in \mathcal{R}^{d \times r}$ followed by an upstream projection $w_t^{\mathrm{up}} \in \mathcal{R}^{r \times u}$, resulting in
\begin{equation}
    \label{eq:adapter}
    a^{a}_{t+1} = w_t^{\mathrm{up}} \cdot h(w_{t}^{\mathrm{down}} \cdot a^{a}_t) \mathrm{.}
\end{equation}

With this, we can fine-tune an FM over much fewer dimensions than when fully fine-tuning a model. This saves both computational resources and communication costs in the same range as FedPEFT.

An improvement with regard to computational and communication efficiency over additional adapters is low-rank adapters (LoRA) \cite{Hu2021}. 
The technique uses a lower dimensional representation $\mathbf{A} \in \mathcal{R}^{r \times u}$, where $u$ is the dimension of the next layer after the LoRA adapter and $\mathbf{B} \in \mathcal{R}^{d \times r}$, where $d$ is the dimension of the previous LoRA adapter. The weight updates are calculated with $w_{t+1} = w_t + \Delta w = w_t + \mathbf{B}  \mathbf{A}$. 
$r \ll \min(d, u)$ casts the weight update matrices into a much lower dimensionality than in the original transformer module without the necessity of adding additional adapters, i.e., LoRA builds an adapter for existing parameters. 
However, as $\mathbf{A}$ is initialized randomly to a Gaussian distribution and $\mathbf{B}$ as a zero matrix, this works well for centralized settings with IID data \cite{Hu2021}. 
For FL settings, this initialization method bears the risk of slowing down the fine-tuning process over non-IID data. 

FedCLIP \cite{Lu2023_fedclip} introduces a PEFT method with an adjusted FedAvg-based adapter aggregation technique. Their approach yields significant performance improvements over vanilla FedAvg and FedProx.

\mycite{Zhang2023} provide a systematic benchmark study on adapter-based fine-tuning methods in privacy-preserving FL systems. Their results show that fine-tuning with additional adapters and LoRA both yield the same benchmark results regarding model accuracy. However, LoRA requires $66\%$ less communication than additional adapters. 

However, both FedCLIP and FedPETuning yield a worse accuracy than full fine-tuning. SLoRA \cite{Babakniya2023} and FedDPA \cite{Yang2024DualPersonalizingAF} are LoRA-based techniques to fine-tune models in non-IID FL settings. Their approach parameterizes the weight update based on $r$, 
\begin{equation}
    \label{eq:slora}
    w_{t+1} = w_t + \frac{\beta}{r} \mathbf{B}  \mathbf{A} \mathrm{.}
\end{equation}

As $\beta$ depends on $r$, scaling the ratio helps control the weight update impact of a single client. Subsequently, this can be used to control inconsistent training updates caused by non-IID data. To practically achieve this, \mycite{Babakniya2023} make use of a two-stage process: First, they used singular vector decomposition on $\mathbf{A}$ and $\mathbf{B}$ to obtain a common initialization point for LoRA across all clients in an FL system. Second, the training is facilitated with the commonly initialized low-rank representations. Their approach achieves on-par performance with full model fine-tuning. However, they require a warmup time of approximately 100 FL rounds for stage 1, which can be very expensive in FL settings as clients are often unavailable consecutively for such a large number of FL rounds \cite{mcmahan2017}.

\subsection{Prompt Tuning}
\label{sec:prompt_tuning}
Prompt tuning is another efficient method for tuning pre-trained models to a downstream task. Here, we use binary sentiments subsequent to masked-language-modeling to achieve high-quality results \cite{Lester2021}. As such, the model remains entirely frozen, and we only tweak the prompts (a very small number of tokens) that are being prepended to each embedded input query to improve the output quality. In contrast to fine-tuning, this method does not interfere with the model architecture or parameters. \mycite{Lester2021} show that the effects of prompt tuning on model performance in centralized settings become better with larger models, i.e., for large FMs, prompt tuning bears significant potential. 

Specifically, in FL settings, for each client $n$ the likelihood $P$ for a desired output $\hat{x}$ is calculated over prepending trainable embedded prompts $x_p$ to each embedded input $x$. In the interactive training process, the prompt is optimized in such a way that it optimally resembles the local objective of a client, 
\begin{equation}
    \label{eq:prompt-tuning}
    \max(P_n(\hat{x}|[x_p;x])),\, \forall n \in N\mathrm{.}
\end{equation}

\mycite{zhao2022} introduce FedPrompt, a method to efficiently communicate federally generated prompts only and aggregate them such that the global model performance of a pre-trained model improves for a downstream task. 
Their experimental evaluation shows a general sensitivity of prompt tuning towards data heterogeneity as the model performance degrades by 5 -- 10\% for the 100M parameter BERT model compared to a centrally trained baseline. 
However, with RoBERTa Base (124M parameters), the sensitivity diminishes, and the FL results are on par with centralized training. 
The larger T5 Base model (223M parameters) follows this trend, showing that prompt tuning becomes more effective with larger model sizes \cite{Lester2021}. 

\subsection{Instruction Tuning}
\label{sec:instruction_tuning}
Some applications work with highly sensitive and protected data or require a very high model performance.
The previously mentioned fine-tuning techniques may not yield sufficient results in these cases.

This is where instruction tuning comes into play as a technique that uses high-quality data. For instance, GPT-2 \cite{radford2019language} uses Reinforcement Learning with Human Feedback (RLHF). RLHF is a multi-stage process where an FM is initially trained on supervised data. In the second step - reward model training - the FM generates outputs over a given prompt, which a user then ranks based on their preference. With this, the model learns human preferences. In the third step - proximal policy optimization - the model trains self-supervised for a maximum reward \cite{Zheng2023_ppo}.

With FedIT, \mycite{zhang2023_fedgpt} study instruction tuning on LLaMA-7B in an FL application over heterogeneous client tasks, e.g., learning brainstorming and text summarization on different clients in a single system at the same time. Their results show that the additional context on a downstream task generated with federated instruction tuning provides net benefits over central training only, even in heterogeneous settings. However, these results were only produced over a single dataset, Dollybricks-15k.

\subsection{Discussion}
As more open-source FMs become available for unrestricted use (e.g., Alpaca \cite{alpaca}, Falcon \cite{falcon_llm}), it is unlikely that full model training will be a common use case since training an FM from scratch is very challenging, even in a centralized setting. 

Therefore, we see a priority in improving upon PEFT for downstream tasks, for instance, by introducing new and enhancing existing algorithms to remove the currently required warm-up times to improve the performance of LoRA in non-IID data environments. Computer Vision (CV) applications may benefit from exploring prompt tuning for vision transformers \cite{Jia2022}.
Also, we find significant challenges for instruction tuning as data quality is a general issue in FL \cite{Longpre2023}, and access to human preferences, as it is required for RLHF, is hardly available in a real-world federated setting without incentive schemes.

Furthermore, PEFT also positively affects communication efficiency by significantly reducing the number of trainable (and thus communicable) parameters. As such, we now see computational and communication efficiency growing closer for FL and FMs, but not to a sufficient degree. As such, there is still a stark need to develop new approaches to use PEFT to improve computational and communication efficiency.

\begin{table*}[!ht]
    \centering
    \scalebox{0.76}{
        \input{tables/Communication_for_FL}
    }
    \caption{Communication-efficient FL methods. Their centralized learning pendants are often tied to specific domains: CV \protect\cite{Habib2023_cv_distil}, NLP \protect\cite{He2021_nlp_distill}, and Audio \protect\cite{Perez_2020_WACV}.}
    \label{tab:summary_communication_works}
\end{table*}

%% file: tables/Communication_for_FL.tex
\begin{tabular}{l|l|l|rr|ccc}
    \toprule
    \multicolumn{1}{l}{}            & \multicolumn{1}{c}{Enhancement}       & \multicolumn{1}{c}{Underlying FL}         & \multicolumn{1}{c}{Max. Model}  & \multicolumn{1}{c}{Communication}               & \multicolumn{3}{c}{Domain} \\
    \multicolumn{1}{l}{Algorithm}   & \multicolumn{1}{c}{Method(s)}         & \multicolumn{1}{c}{Aggregation Strategy}  & \multicolumn{1}{c}{Parameters}  & \multicolumn{1}{c}{Savings vs. FedAvg}          & CV & NLP & Audio \\                 
    \midrule
    \textit{Centralized Learning}   & MP, Q, S, GP                         & --                                        & $\geq 7\mathrm{B}$               & --                                              & \checkmark    & \checkmark    & \checkmark  \\
    \midrule                                                                                                   
    FedKSeed \cite{Qin2023} & GP                                   & Weighted Average                          & 3B                              & $\geq 99\%$                                     &               & \checkmark & \\   
    FedOBD \cite{Chen2023}         & Q                                      & Weighted Average                          & 17M                             & $89\%$                                          & \checkmark    & \checkmark & \\
    PruneFL \cite{Jiang2022}       & MP                                     & Weighted Average                          & 132M                            & $80\%$                                          & \checkmark    &   &   \\
    FedPM \cite{Isik2022}          & MP                                     & Weighted Average                          & 12M                             & $98\%$                                          & \checkmark    &   &   \\
    FedTiny \cite{Huang2022}       & MP                                     & Weighted Average                          & 132M                            & $97\%$                                          &\checkmark     &   &   \\
    SoteriaFL \cite{Li2022}        & S                                      & FedSGD                                    & 0.05M                           & N/A                                             & \checkmark    &   &   \\
    FjORD \cite{horvath2021fjord}  & MP                                     & Weighted Average                          & 11M                             & $98\%$                                          & \checkmark    & \checkmark & \\
    FedPAQ \cite{Reisizadeh2020}   & Q                                      & FedSGD                                    & 0.2M                           & $90\%$                                          & \checkmark    &   &   \\
    HeteroFL \cite{Diao2020}       & MP                                     & Weighted Average                          & 11M                             & $98\%$                                          & \checkmark    & & \\
    LotteryFL \cite{Ang2020}       & MP                                     & Weighted Average                          & 138M                            & $50\%$                                          & \checkmark    & & \\
                                              
    \bottomrule                                
    \bottomrule
\end{tabular}

%% file: chapters/04_importance_of_communication.tex
With FMs, models have exponentially grown from a few million to several billion parameters to be able to serve multi-modal tasks \cite{bommasani2021}. 
For FL, this specifically means that the communication load between clients and servers has grown significantly, even though with PEFT, there is not necessarily the need to communicate an entire model. 
However, when fine-tuning a billion-parameter FM with adapter-based methods, we still need to facilitate communication for millions of parameters. 
For cross-device scenarios involving more than 1,000 clients per training round, the data traffic can quickly overstrain a server's network capacity and potentially incur significant communication costs for data transfer from the edge to a cloud \cite{Xu2023-qg,Isenko2023}.
Therefore, efficient communication and training design is vital for future FL systems. We distinguish two major efficiency methods: model pruning and full model compression. A detailed overview of studies on efficient communication in FL is provided in \Cref{tab:summary_communication_works}.

\subsection{Model Pruning}
\label{sec:pruning}
The objective of model pruning (MP) is to retain and communicate only parts of a DL model that are relevant to a certain task. 
The reduction of parameters with this technique reduces the communication effort \cite{zhu2017_prune}. However, the success of pruning highly depends on the underlying data. 
Pruning client models without coordination may deny convergence with heterogeneous non-IID data in FL systems. 
\mycite{Jiang2022} introduce PruneFL, a two-stage procedure to realize model pruning in FL systems. 
The first stage is carried out on a powerful client to find a common initialization for the model and generate the importance-based pruning mask. 
This mask is then iteratively refined over multiple FL training rounds under the consideration of all clients. 
The experimental results with PruneFL show two remarkable results: (I) The models have a shorter time to accuracy over the same task with PruneFL than with FedAvg, which is attributable to the higher degree of model specialization. 
(II) Since the model size is reduced, one would expect additional effects on faster computation, but there is limited hardware support for sparse matrix multiplications in training as they are required in PruneFL. 
Therefore, PruneFL has no computational benefits to this point. However, this may change with new hardware, such as sparse Tensor cores that support PruneFL's dynamic pruning approach \cite{Zhu2019_sparse}. 
With FedTiny, \mycite{Huang2022} present an approach that works identically to PruneFL, except for them swapping the common initialization procedure with using batch normalization values of clients to choose a common initialization. 
Furthermore, FjORD \cite{horvath2021fjord} and HeteroFL \cite{Diao2020} provide similar approaches to pruning.

\mycite{Isik2022} choose a similar approach for pruning models based on the lottery ticket hypothesis, first introduced to FL by \mycite{Ang2020}. Instead of commonly initializing a pruned model for an FL system, they initialize a random binary mask based on a shared seed on each client. This reduces computational efforts in the ramp-up phase. After an FL training round, each client communicates their binary mask to the server, which creates a global model based on the weighted average of those binary masks. With this, an approximate weight estimate replaces the parameters on the global model. From client to server, FedPM achieves significant communication efficiencies. However, the full model still has to be communicated from the server to the clients, lowering the net benefit.

Model pruning has also been discussed extensively for FM fine-tuning outside of FL \cite{lagunas-etal-2021-block,sanh2020}.
As existing pruning approaches have shown strong benefits to delivering on-par model performance compared to fine-tuning the full model, this is a promising direction to combine federated PEFT with highly efficient pruning techniques to further enhance communication. Along with pruning, sparse tensor hardware can  lower computational loads.

\subsection{Full Model Compression}
\label{sec:full_model_compression}
Model pruning is prone to omit segments of a DL model that may become relevant at a later stage. This originates from domain shifts \cite{Peng2020Federated_domainshift} and might require preserving the full model with all its parameters. For this, three frequently discussed techniques for full model compression in FL systems are Quantization, Sparsification, and Gradient Projection.

\paragraph{Quantization (Q).} The first work towards dynamic quantization is FedPAQ \cite{Reisizadeh2020}, which combines FedAvg with strong quantization guarantees, where $Q$ represents the quantization term for a local model update, 
\begin{equation}
    \label{eq:quantization}
    w_{t+1} = w_t + \frac{1}{|N|} \sum^{|N|}_{n = 1} Q(w^{n}_{t+1} - w_t) \mathrm{.}
\end{equation}

While DL models often operate on full precision (32-bit), this high degree of detail is not necessarily required \cite{Zhou2018_quantization}. 
FedPAQ leverages this to reduce the communication intensity of FL applications. 
$Q$ calculates the optimal float precision of a model update to preserve all required information: $Q(w) = \norm{w} \cdot \mathrm{sign}(w) \cdot \xi(w, s)$, as proposed in QSGD by \mycite{Alistarh2017}. 
$\xi$ formulates a stochastic process to dynamically tune $s$, the level of precision. FedPAQ has a significantly lower time to accuracy than QSGD, which is attributable to dynamizing $s$. However, it must be noted that dynamic quantization only yields benefits for communication. Depending on the infrastructure, the model updates have to be cast back to full precision, creating additional computational overhead on the client and server. Also, the method has been only tested with small models ($< 100\mathrm{K}$ parameters). 

FedOBD \cite{Chen2023} quantizes models with transformer block dropout, i.e., the random removal of entire model blocks. The dropout mechanism is carried out during training by each client and returns only the top-k most important model blocks to communicate. Additionally, FedOBD includes the ideas discussed in \mycite{Alistarh2017} and \mycite{Reisizadeh2020} but proposes an optimization problem out of the stochastic quantization where the trade-off originates from entropy and update size. The communication required for FL with a 17M parameter transformer model shows FedOBD to cut communication cost by $2\times$ vs. FedPAQ and by $8\times$ compared to vanilla FedAvg \cite{Chen2023}.

\paragraph{Sparsification (S).} While model pruning and sparsification technically have the same objective, pruned models do not necessarily resemble sparse models. A model is sparse once more than 50\% of weights are set to $0$ \cite{frankle2018_sparse}. However, pruning can also change the model architecture, i.e., not return the full model. Since it lends its idea from \cite{frankle2018_sparse}, FedPM \cite{Isik2022} (see \Cref{sec:pruning}) can be considered as a model sparsification technique but does not necessarily lead to sparse networks and may return partial networks. 
SoteriaFL \cite{Li2022} guarantees sparse networks while maintaining differential privacy. \Cref{eq:quantization} is amended in such a way that $Q(w^n_{t+1} - w_t)$ is replaced by $C(w^n_{t+1} + \mathcal{N}(0, \sigma^2\mathcal{I}))$ with $C$ resembling a sparse client update through shifted compression that has proven to improve convergence of DL models in FL settings compared to direct compression \cite{mitra2021linear,Mishchenko2019_sparse}. Overall, SoteriaFL mitigates the trade-off between model utility and compression, i.e., the differentially private models converge faster in stricter compression regimes than previously existing non-compressed differentially private approaches.  

\paragraph{Gradient projection (GP).} 
\mycite{Qin2023} introduce FedKSeed tailored to efficiently train FMs in FL applications. They do so by using seeds in the form of scalar vectors to create gradient projections. As only the scalar vectors have to be transmitted, the total amount of communication is reduced by $\geq 99\%$ compared to applications that would send the original multi-dimensional gradients.
This is the first technique that enables communication efficient FL applications with FMs, regardless of the training regime (pre-training or fine-tuning).   


\subsection{Discussion}
To date, advancements in communication-efficient methods for FL systems have predominantly focused on training small, full models. 
The communication paradigm shifts with the emergence of FMs in FL applications. 
With fine-tuning tasks, we only need to train a small fraction of model parameters, and thus, only trainable parameters have to be communicated. 
However, each trainable parameter usually contains a high degree of information for a downstream task. As such, the effectiveness of model pruning techniques is unclear as they would cut away fine-tuned parameters.
An unexplored space in FL research is the pruning of FMs with subsequent fine-tuning for downstream tasks. Pruning FMs can lead to smaller transformer layers and, consequently, smaller PEFT layers with fewer parameters. In turn, this could positively affect computational and communication efficiency.
Full-model compression techniques do not alter the model structure but rather reduce the parameter precision. 
Thus, these techniques can be used with FL applications and FMs in order to further reduce the size of the communicated updates. 
Furthermore, \mycite{Qin2023} have shown that gradient projection is a promising direction to compress model updates without sacrificing significant information. This can be beneficial for PEFT applications as we have small specific adapters for downstream tasks. However, the remaining key challenge is the effect of non-IID data on PEFT. Potential compounding effects of communication compression and lossy compression remain open for investigation.

%% file: chapters/05_fl_systems_with_fm_and_llm.tex
The backbones for making FL applications available to a broad audience are FL frameworks implementing recent advancements in FL research.
We investigate widely used frameworks for their FM readiness and progress in integrating computational and communication efficiency (\Cref{tab:sota_fl_systems}). 

\begin{table}
    \centering
    \resizebox{0.49\textwidth}{!}{
        \input{tables/fl_system_status}
    }
    \caption{Current capabilities of state-of-the-art FL framework with respect to our taxonomy and their ability to run in resource-limited environments. {Key: DP = Differential Privacy, HEC = Homomorphic Encryption, SMPC = Secure Multi-Party Computation.}}
    \label{tab:sota_fl_systems}
\end{table}

In theory, all FL frameworks could handle FMs with sufficient hardware availability. 
Yet, only some implement efficient training methods. 
To further drive the adoption of FL in times of FMs, the frameworks need to improve both computationally- and communication-efficient methods for training.
FL frameworks that are characterized by their active open-source community, FLARE \cite{Roth2022}, FATE \cite{Fan2023_fate_llm,Liu2021_fate}, FedML \cite{He2020_fedml}, TensorFlow Federated (TFF) \cite{tensorflow_federated}, FederatedScope \cite{Kuang_2023_fscope_llm,Xie2023_fscope} and Flower \cite{beutel2020_flower}, have adopted recent advancements in FL research. The frameworks especially allow for PEFT of FMs with LoRA.
Substra \cite{Galtier2019_substra}, PySyft \cite{Ziller2021_pysyft}, OpenFL \cite{Foley2022_openfl}, and IBM FL \cite{Ludwig2020_ibmfl}, in their versions as of 2023, focus on training smaller FL tasks with 100K up to a 10M parameters and, therefore, do not provide adapters for FM workloads with more than 100M parameters.
Yet, a consistent observation across all frameworks is their lack of efficient communication techniques (e.g., FedOBD). 
Workloads with FMs will significantly increase communication costs, and the growing use cases involving resource-constrained edge and IoT devices require high efficiency for computation and communication. 
As such, only those FL frameworks enabling training efficiency are viable choices for working with FMs.

%% file: tables/fl_system_status.tex
\begin{tabular}{lllccc}
    \toprule
                       & \multicolumn{1}{c}{\textbf{Secure}}        & \multicolumn{1}{c}{\textbf{Training}}     & \multicolumn{1}{c}{\textbf{Communication}} & \textbf{FM Training /}       & \textbf{Edge}             \\
    \textbf{Framework} & \multicolumn{1}{c}{\textbf{Aggregation}}   & \multicolumn{1}{c}{\textbf{Efficiency}}   & \multicolumn{1}{c}{\textbf{Efficiency}}    & \textbf{Fine-Tuning}         & \textbf{Ready}            \\
    \midrule
    FLARE              & DP, HEC                                    &                                           & \multirow{10}{*}{\shortstack[l]{Currently, none of the \\ Frameworks implements \\ communication efficient \\ FL methods.}}                       & \checkmark                   &                           \\
    FedML              & DP, HEC                                    & PEFT                                      &                                           & \checkmark                   & \checkmark                \\
    FederatedScope     &                                            & PEFT                                      &                                           & \checkmark                   & \checkmark                \\
    Flower             & DP, SMPC                                   & PEFT                                      &                                           & \checkmark                   & \checkmark                \\
    FATE               & HEC, SMPC                                  & PEFT                                      &                                           & \checkmark                   & \checkmark                \\ 
    \cmidrule{1-3}\cmidrule{5-6}
    Substra            &                                            &                                           &                                           &                              &                           \\
    PySyft             & DP                                         &                                           &                                           &                              &                           \\
    OpenFL             &                                            &                                           &                                           &                              &                           \\
    TFF                & DP                                         &                                           &                                           &                              & \checkmark                \\
    IBM FL             & HEC                                        &                                           &                                           &                              &                           \\
    \bottomrule
    \bottomrule
\end{tabular}

%% file: chapters/06_related_work.tex
While there are ample surveys that provide a broad perspective on FL \cite{Li2023,Banabilah2022,Liu2022,Nguyen2021,Zhang2021,Aledhari2020}, there are two closely related surveys to our work as they also focus on FMs.

In their survey, \mycite{zhuang2023} introduce a broad and general perspective on FMs and FL. 
They extensively discuss data modalities. This includes access to data across a large number of highly distributed clients and the quality of data that lives on these clients.
Currently, FM training or fine-tuning requires datasets with high data quality, i.e., the instructions or texts used for MLM must be curated very carefully. 
Thus, their survey identifies a stark need for methods to train or fine-tune FMs on a scattered data basis with (highly) varying data quality. 
Further, \mycite{zhuang2023} discuss approaches to integrate FL applications into the lifecycle of FMs, i.e., how FMs can benefit from a continuously evolving system.
While their survey briefly touches upon computational efficiency, our study provides an in-depth overview of state-of-the-art training and fine-tuning techniques to render FMs in FL applications a reality. 
Furthermore, our study includes a comprehensive overview of communication techniques that can enhance the adoption of FMs in communication resource-constrained environments.

\mycite{yu2023_fl_llm} provide an overview of FMs and FL with a special focus on privacy, an integral component of FL. 
Their survey includes a comprehensive overview of different fields of application for FMs, which they divide into once-off training and continual learning. 
The authors elaborate on technical challenges that may arise for specific use cases, such as robustness towards unreliable clients, varying data quality, the degree of non-IID data, and scalability. 
In contrast, our survey provides an application-agnostic, in-depth study of existing methods suitable for FM training. Our focus is to outline the technical challenges that currently hinder the operationalization of FM for use cases in federated applications.

%% file: chapters/07_conclusion.tex
In this paper, we survey the current landscape of computational and communication efficiency methods in FL systems and introduce a novel taxonomy based on the key techniques.
While efficient FL methods have been separate topics on their own in the past, they become closely intertwined as we start using FL systems to leverage FMs. Consequently, the following three questions arise:

\paragraph{What are good and realistic evaluation strategies for generative downstream tasks in FL settings where we do not have control of data?}
Fine-tuning generative FM requires high-quality data. 
However, we do not have access to data on the clients to monitor data quality before or during training.
As such, estimating data quality is of utmost importance.

\paragraph{How does hyperparameter optimization work for FMs in continuously evolving FL systems?} 
While hyperparameter optimization in FL has been a key challenge, PEFT adds additional complexity. 
As such FL systems must adapt to the era of FMs by introducing adaptive parameterization for PEFT techniques that can account for changing environmental conditions.


\paragraph{We must develop an understanding of the interplay between PEFT and privacy in FL systems.}
Communication-efficient FL techniques have been studied for their effect on privacy, but this is still an open topic for PEFT, PT, and IT. While it is proven that PEFT is more sensitive to data heterogeneity, the effects of perturbation through differential privacy are still subject to further studies. The same is true for PT and IT, as both techniques require precise prompts and instructions, respectively. As such, noise may have significantly negative effects here, as well.

%% file: ijcai24.bbl
\begin{thebibliography}{}

\bibitem[\protect\citeauthoryear{Aledhari \bgroup \em et al.\egroup }{2020}]{Aledhari2020}
Mohammed Aledhari, Rehma Razzak, Reza~M. Parizi, and Fahad Saeed.
\newblock Federated learning: A survey on enabling technologies, protocols, and applications.
\newblock {\em IEEE Access}, 8:140699–140725, 2020.

\bibitem[\protect\citeauthoryear{Alistarh \bgroup \em et al.\egroup }{2017}]{Alistarh2017}
Dan Alistarh, Demjan Grubic, et~al.
\newblock Qsgd: Communication-efficient sgd via gradient quantization and encoding.
\newblock In {\em Advances in Neural Information Processing Systems}, volume~30, 2017.

\bibitem[\protect\citeauthoryear{Babakniya \bgroup \em et al.\egroup }{2023}]{Babakniya2023}
Sara Babakniya, Ahmed~R. Elkordy, et~al.
\newblock Slora: Federated parameter efficient fine-tuning of language models, 2023.

\bibitem[\protect\citeauthoryear{Banabilah \bgroup \em et al.\egroup }{2022}]{Banabilah2022}
Syreen Banabilah, Moayad Aloqaily, et~al.
\newblock Federated learning review: Fundamentals, enabling technologies, and future applications.
\newblock {\em Information Processing \& Management}, 59(6), 2022.

\bibitem[\protect\citeauthoryear{Beutel \bgroup \em et al.\egroup }{2020}]{beutel2020_flower}
Daniel~J. Beutel, Taner Topal, et~al.
\newblock Flower: A friendly federated learning research framework, 2020.

\bibitem[\protect\citeauthoryear{Bommasani \bgroup \em et al.\egroup }{2021}]{bommasani2021}
Rishi Bommasani, Drew~A. Hudson, et~al.
\newblock On the opportunities and risks of foundation models, 2021.

\bibitem[\protect\citeauthoryear{Chen \bgroup \em et al.\egroup }{2023}]{Chen2023}
Yuanyuan Chen, Zichen Chen, et~al.
\newblock Fedobd: Opportunistic block dropout for efficiently training large-scale neural networks through federated learning.
\newblock In {\em Proceedings of the Thirty-Second International Joint Conference on Artificial Intelligence, {IJCAI-23}}. IJCAI Org., 2023.

\bibitem[\protect\citeauthoryear{Diao \bgroup \em et al.\egroup }{2020}]{Diao2020}
Enmao Diao, Jie Ding, et~al.
\newblock Heterofl: Computation and communication efficient federated learning for heterogeneous clients, 2020.

\bibitem[\protect\citeauthoryear{Ding \bgroup \em et al.\egroup }{2023}]{Ding2023_nature}
Ning Ding, Yujia Qin, et~al.
\newblock Parameter-efficient fine-tuning of large-scale pre-trained language models.
\newblock {\em Nature Machine Intelligence}, 5(3):220–235, March 2023.

\bibitem[\protect\citeauthoryear{Dosovitskiy \bgroup \em et al.\egroup }{2020}]{Dosovitskiy2020}
Alexey Dosovitskiy, Lucas Beyer, et~al.
\newblock An image is worth 16x16 words: Transformers for image recognition at scale, 2020.

\bibitem[\protect\citeauthoryear{Erben \bgroup \em et al.\egroup }{2023}]{Isenko2023}
Alexander Erben, Ruben Mayer, and Hans-Arno Jacobsen.
\newblock How can we train deep learning models across clouds and continents? an experimental study, 2023.

\bibitem[\protect\citeauthoryear{Fan \bgroup \em et al.\egroup }{2023}]{Fan2023_fate_llm}
Tao Fan, Yan Kang, et~al.
\newblock Fate-llm: A industrial grade federated learning framework for large language models, 2023.

\bibitem[\protect\citeauthoryear{Foley \bgroup \em et al.\egroup }{2022}]{Foley2022_openfl}
Patrick Foley, Micah~J Sheller, et~al.
\newblock Openfl: the open federated learning library.
\newblock {\em Physics in Medicine \& Biology}, 67(21), 2022.

\bibitem[\protect\citeauthoryear{Frankle and Carbin}{2019}]{frankle2018_sparse}
Jonathan Frankle and Michael Carbin.
\newblock The lottery ticket hypothesis: Finding sparse, trainable neural networks.
\newblock In {\em ICLR}, 2019.

\bibitem[\protect\citeauthoryear{Galtier and Marini}{2019}]{Galtier2019_substra}
Mathieu~N Galtier and Camille Marini.
\newblock Substra: a framework for privacy-preserving, traceable and collaborative machine learning, 2019.

\bibitem[\protect\citeauthoryear{Google}{2019}]{tensorflow_federated}
Inc. Google.
\newblock Tensorflow federated: Machine learning on decentralized data.
\newblock \url{https://www.tensorflow.org/federated}, 2019.
\newblock Accessed: 2023-12-01.

\bibitem[\protect\citeauthoryear{Habib \bgroup \em et al.\egroup }{2023}]{Habib2023_cv_distil}
Gousia Habib, Tausifa~Jan Saleem, and Brejesh Lall.
\newblock Knowledge distillation in vision transformers: A critical review, 2023.

\bibitem[\protect\citeauthoryear{He \bgroup \em et al.\egroup }{2020}]{He2020_fedml}
Chaoyang He, Songze Li, et~al.
\newblock Fedml: A research library and benchmark for federated machine learning, 2020.

\bibitem[\protect\citeauthoryear{He \bgroup \em et al.\egroup }{2021}]{He2021_nlp_distill}
Haoyu He, Xingjian Shi, et~al.
\newblock Distiller: A systematic study of model distillation methods in natural language processing, 2021.

\bibitem[\protect\citeauthoryear{Horv{\'a}th \bgroup \em et al.\egroup }{2021}]{horvath2021fjord}
Samuel Horv{\'a}th, Stefanos Laskaridis, et~al.
\newblock Fj{ORD}: Fair and accurate federated learning under heterogeneous targets with ordered dropout.
\newblock In {\em Advances in Neural Information Processing Systems}, 2021.

\bibitem[\protect\citeauthoryear{Houlsby \bgroup \em et al.\egroup }{2019}]{Houlsby2019}
Neil Houlsby, Andrei Giurgiu, et~al.
\newblock Parameter-efficient transfer learning for nlp, 2019.

\bibitem[\protect\citeauthoryear{Hu \bgroup \em et al.\egroup }{2021}]{Hu2021}
Edward~J. Hu, Yelong Shen, et~al.
\newblock Lora: Low-rank adaptation of large language models, 2021.

\bibitem[\protect\citeauthoryear{Huang \bgroup \em et al.\egroup }{2022}]{Huang2022}
Hong Huang, Lan Zhang, et~al.
\newblock Distributed pruning towards tiny neural networks in federated learning, 2022.

\bibitem[\protect\citeauthoryear{Isik \bgroup \em et al.\egroup }{2022}]{Isik2022}
Berivan Isik, Francesco Pase, et~al.
\newblock Sparse random networks for communication-efficient federated learning, 2022.

\bibitem[\protect\citeauthoryear{Jia \bgroup \em et al.\egroup }{2022}]{Jia2022}
Menglin Jia, Luming Tang, et~al.
\newblock Visual prompt tuning, 2022.

\bibitem[\protect\citeauthoryear{Jiang \bgroup \em et al.\egroup }{2022}]{Jiang2022}
Yuang Jiang, Shiqiang Wang, et~al.
\newblock Model pruning enables efficient federated learning on edge devices.
\newblock {\em {IEEE} Transactions on Neural Networks and Learning Systems}, 2022.

\bibitem[\protect\citeauthoryear{Kuang \bgroup \em et al.\egroup }{2023}]{Kuang_2023_fscope_llm}
Weirui Kuang, Bingchen Qian, et~al.
\newblock Federatedscope-llm: A comprehensive package for fine-tuning large language models in federated learning, 2023.

\bibitem[\protect\citeauthoryear{Lagunas \bgroup \em et al.\egroup }{2021}]{lagunas-etal-2021-block}
Francois Lagunas, Ella Charlaix, et~al.
\newblock Block pruning for faster transformers.
\newblock In {\em Proceedings of the 2021 Conference on EMNLP}, Online and Punta Cana, Dominican Republic, November 2021. ACL.

\bibitem[\protect\citeauthoryear{Lester \bgroup \em et al.\egroup }{2021}]{Lester2021}
Brian Lester, Rami Al-Rfou, and Noah Constant.
\newblock The power of scale for parameter-efficient prompt tuning, 2021.

\bibitem[\protect\citeauthoryear{Li \bgroup \em et al.\egroup }{2020}]{Ang2020}
Ang Li, Jingwei Sun, et~al.
\newblock Lotteryfl: Personalized and communication-efficient federated learning with lottery ticket hypothesis on non-iid datasets, 2020.

\bibitem[\protect\citeauthoryear{Li \bgroup \em et al.\egroup }{2022}]{Li2022}
Zhize Li, Haoyu Zhao, et~al.
\newblock Soteriafl: A unified framework for private federated learning with communication compression.
\newblock In {\em Advances in Neural Information Processing Systems}, volume~35, 2022.

\bibitem[\protect\citeauthoryear{Li \bgroup \em et al.\egroup }{2023}]{Li2023}
Qinbin Li, Zeyi Wen, Zhaomin Wu, Sixu Hu, Naibo Wang, Yuan Li, Xu~Liu, and Bingsheng He.
\newblock A survey on federated learning systems: Vision, hype and reality for data privacy and protection.
\newblock {\em IEEE Transactions on Knowledge and Data Engineering}, 35(4):3347–3366, April 2023.

\bibitem[\protect\citeauthoryear{Liu \bgroup \em et al.\egroup }{2021}]{Liu2021_fate}
Yang Liu, Tao Fan, et~al.
\newblock Fate: An industrial grade platform for collaborative learning with data protection.
\newblock {\em JMLR}, 22(1), 2021.

\bibitem[\protect\citeauthoryear{Liu \bgroup \em et al.\egroup }{2022}]{Liu2022}
Ji~Liu, Jizhou Huang, et~al.
\newblock From distributed machine learning to federated learning: a survey.
\newblock {\em Knowledge and Information Systems}, 64(4), 2022.

\bibitem[\protect\citeauthoryear{Longpre \bgroup \em et al.\egroup }{2023}]{Longpre2023}
Shayne Longpre, Le~Hou, et~al.
\newblock The flan collection: Designing data and methods for effective instruction tuning, 2023.

\bibitem[\protect\citeauthoryear{Lu \bgroup \em et al.\egroup }{2023}]{Lu2023_fedclip}
Wang Lu, Xixu Hu, et~al.
\newblock Fedclip: Fast generalization and personalization for clip in federated learning, 2023.

\bibitem[\protect\citeauthoryear{Ludwig \bgroup \em et al.\egroup }{2020}]{Ludwig2020_ibmfl}
Heiko Ludwig, Nathalie Baracaldo, et~al.
\newblock Ibm federated learning: an enterprise framework white paper v0.1, 2020.

\bibitem[\protect\citeauthoryear{McMahan \bgroup \em et al.\egroup }{2017}]{mcmahan2017}
Brendan McMahan, Eider Moore, et~al.
\newblock {Communication-Efficient Learning of Deep Networks from Decentralized Data}.
\newblock In {\em Proceedings of the 20th International Conference on Artificial Intelligence and Statistics}, volume~54 of {\em Proceedings of Machine Learning Research}. PMLR, 2017.

\bibitem[\protect\citeauthoryear{Mishchenko \bgroup \em et al.\egroup }{2019}]{Mishchenko2019_sparse}
Konstantin Mishchenko, Eduard Gorbunov, et~al.
\newblock Distributed learning with compressed gradient differences, 2019.

\bibitem[\protect\citeauthoryear{Mitra \bgroup \em et al.\egroup }{2021}]{mitra2021linear}
Aritra Mitra, Rayana Jaafar, George~J. Pappas, and Hamed Hassani.
\newblock Linear convergence in federated learning: Tackling client heterogeneity and sparse gradients.
\newblock In {\em Advances in Neural Information Processing Systems}, 2021.

\bibitem[\protect\citeauthoryear{Nguyen \bgroup \em et al.\egroup }{2021}]{Nguyen2021}
Dinh~C. Nguyen, Ming Ding, et~al.
\newblock Federated learning for internet of things: A comprehensive survey.
\newblock {\em IEEE Communications Surveys \& Tutorials}, 23(3), 2021.

\bibitem[\protect\citeauthoryear{OpenAI}{2023}]{gpt4_paper}
OpenAI.
\newblock {GPT-4} technical report, 2023.

\bibitem[\protect\citeauthoryear{Penedo \bgroup \em et al.\egroup }{2023}]{falcon_llm}
Guilherme Penedo, Quentin Malartic, et~al.
\newblock The refinedweb dataset for falcon llm: Outperforming curated corpora with web data, and web data only, 2023.

\bibitem[\protect\citeauthoryear{Peng \bgroup \em et al.\egroup }{2020}]{Peng2020Federated_domainshift}
Xingchao Peng, Zijun Huang, et~al.
\newblock Federated adversarial domain adaptation.
\newblock In {\em ICLR}, 2020.

\bibitem[\protect\citeauthoryear{Perez \bgroup \em et al.\egroup }{2020}]{Perez_2020_WACV}
Andres Perez, Valentina Sanguineti, et~al.
\newblock Audio-visual model distillation using acoustic images.
\newblock In {\em Proceedings of the IEEE/CVF Winter Conference on Applications of Computer Vision (WACV)}, 2020.

\bibitem[\protect\citeauthoryear{Qin \bgroup \em et al.\egroup }{2023}]{Qin2023}
Zhen Qin, Daoyuan Chen, et~al.
\newblock Federated full-parameter tuning of billion-sized language models with communication cost under 18 kilobytes, 2023.

\bibitem[\protect\citeauthoryear{Radford \bgroup \em et al.\egroup }{2019}]{radford2019language}
Alec Radford, Jeff Wu, et~al.
\newblock Language models are unsupervised multitask learners.
\newblock 2019.

\bibitem[\protect\citeauthoryear{Reisizadeh \bgroup \em et al.\egroup }{2020}]{Reisizadeh2020}
Amirhossein Reisizadeh, Aryan Mokhtari, et~al.
\newblock Fedpaq: A communication-efficient federated learning method with periodic averaging and quantization.
\newblock In {\em Proceedings of the Twenty Third International Conference on Artificial Intelligence and Statistics}, volume 108. PMLR, 2020.

\bibitem[\protect\citeauthoryear{Roth \bgroup \em et al.\egroup }{2022}]{Roth2022}
Holger~R. Roth, Yan Cheng, et~al.
\newblock Nvidia flare: Federated learning from simulation to real-world.
\newblock 2022.

\bibitem[\protect\citeauthoryear{Ryabinin \bgroup \em et al.\egroup }{2023}]{ryabinin2023}
Max Ryabinin, Tim Dettmers, et~al.
\newblock Swarm parallelism: Training large models can be surprisingly communication-efficient, 2023.

\bibitem[\protect\citeauthoryear{Sanh \bgroup \em et al.\egroup }{2020}]{sanh2020}
Victor Sanh, Thomas Wolf, and Alexander Rush.
\newblock Movement pruning: Adaptive sparsity by fine-tuning.
\newblock In {\em Advances in Neural Information Processing Systems}, volume~33. Curran Associates, Inc., 2020.

\bibitem[\protect\citeauthoryear{Sun \bgroup \em et al.\egroup }{2023}]{sun2023}
Guangyu Sun, Matias Mendieta, et~al.
\newblock Exploring parameter-efficient fine-tuning for improving communication efficiency in federated learning, 2023.

\bibitem[\protect\citeauthoryear{Taori \bgroup \em et al.\egroup }{2023}]{alpaca}
Rohan Taori, Ishaan Gulrajani, et~al.
\newblock Stanford alpaca: An instruction-following llama model.
\newblock \url{https://github.com/tatsu-lab/stanford_alpaca}, 2023.

\bibitem[\protect\citeauthoryear{Tian \bgroup \em et al.\egroup }{2022}]{Tian2022}
Yuanyishu Tian, Yao Wan, et~al.
\newblock Fedbert: When federated learning meets pre-training.
\newblock {\em ACM Transactions on Intelligent Systems and Technology}, 13(4), 2022.

\bibitem[\protect\citeauthoryear{Woisetschl\"{a}ger \bgroup \em et al.\egroup }{2023}]{woisetschlaeger2023}
Herbert Woisetschl\"{a}ger, Alexander Erben, et~al.
\newblock Federated fine-tuning of llms on the very edge: The good, the bad, the ugly, 2023.

\bibitem[\protect\citeauthoryear{Xie \bgroup \em et al.\egroup }{2023}]{Xie2023_fscope}
Yuexiang Xie, Zhen Wang, et~al.
\newblock Federatedscope: A flexible federated learning platform for heterogeneity.
\newblock {\em Proceedings of the VLDB Endowment}, 16(5), 2023.

\bibitem[\protect\citeauthoryear{Xu \bgroup \em et al.\egroup }{2023}]{Xu2023-qg}
Zheng Xu, Yanxiang Zhang, et~al.
\newblock Federated learning of gboard language models with differential privacy.
\newblock 2023.

\bibitem[\protect\citeauthoryear{Yang \bgroup \em et al.\egroup }{2023}]{yan2023_uniaudio}
Dongchao Yang, Jinchuan Tian, et~al.
\newblock Uniaudio: An audio foundation model toward universal audio generation, 2023.

\bibitem[\protect\citeauthoryear{Yang \bgroup \em et al.\egroup }{2024}]{Yang2024DualPersonalizingAF}
Yiyuan Yang, Guodong Long, Taoshu Shen, Jing Jiang, and Michael Blumenstein.
\newblock Dual-personalizing adapter for federated foundation models.
\newblock {\em ArXiv}, abs/2403.19211, 2024.

\bibitem[\protect\citeauthoryear{Yousefpour \bgroup \em et al.\egroup }{2023}]{Yousefpour2023_green_fl}
Ashkan Yousefpour, Shen Guo, et~al.
\newblock Green federated learning, 2023.

\bibitem[\protect\citeauthoryear{Yu \bgroup \em et al.\egroup }{2023}]{yu2023_fl_llm}
Sixing Yu, J.~Pablo Muñoz, and Ali Jannesari.
\newblock Federated foundation models: Privacy-preserving and collaborative learning for large models, 2023.

\bibitem[\protect\citeauthoryear{Zaken \bgroup \em et al.\egroup }{2021}]{zaken_2021_bitfit}
Elad~Ben Zaken, Shauli Ravfogel, and Yoav Goldberg.
\newblock Bitfit: Simple parameter-efficient fine-tuning for transformer-based masked language-models, 2021.

\bibitem[\protect\citeauthoryear{Zhang \bgroup \em et al.\egroup }{2021}]{Zhang2021}
Chen Zhang, Yu~Xie, et~al.
\newblock A survey on federated learning.
\newblock {\em Knowledge-Based Systems}, 216, 2021.

\bibitem[\protect\citeauthoryear{Zhang \bgroup \em et al.\egroup }{2023a}]{zhang2023_fedgpt}
Jianyi Zhang, Saeed Vahidian, et~al.
\newblock Towards building the federated gpt: Federated instruction tuning, 2023.

\bibitem[\protect\citeauthoryear{Zhang \bgroup \em et al.\egroup }{2023b}]{Zhang2023}
Zhuo Zhang, Yuanhang Yang, et~al.
\newblock {FedPETuning}: When federated learning meets the parameter-efficient tuning methods of pre-trained language models.
\newblock In {\em Findings of the Association for Computational Linguistics: {ACL} 2023}. ACL, 2023.

\bibitem[\protect\citeauthoryear{Zhao \bgroup \em et al.\egroup }{2022}]{zhao2022}
Haodong Zhao, Wei Du, et~al.
\newblock Fedprompt: Communication-efficient and privacy preserving prompt tuning in federated learning, 2022.

\bibitem[\protect\citeauthoryear{Zheng \bgroup \em et al.\egroup }{2023}]{Zheng2023_ppo}
Rui Zheng, Shihan Dou, et~al.
\newblock Secrets of rlhf in large language models part i: Ppo, 2023.

\bibitem[\protect\citeauthoryear{Zhou \bgroup \em et al.\egroup }{2018}]{Zhou2018_quantization}
Yiren Zhou, Seyed Moosavi-Dezfooli, et~al.
\newblock Adaptive quantization for deep neural network.
\newblock {\em Proceedings of the AAAI Conference on Artificial Intelligence}, 32(1), 2018.

\bibitem[\protect\citeauthoryear{Zhu and Gupta}{2017}]{zhu2017_prune}
Michael Zhu and Suyog Gupta.
\newblock To prune, or not to prune: exploring the efficacy of pruning for model compression, 2017.

\bibitem[\protect\citeauthoryear{Zhu \bgroup \em et al.\egroup }{2019}]{Zhu2019_sparse}
Maohua Zhu, Tao Zhang, et~al.
\newblock Sparse tensor core: Algorithm and hardware co-design for vector-wise sparse neural networks on modern gpus.
\newblock In {\em Proceedings of the 52nd Annual IEEE/ACM International Symposium on Microarchitecture}, MICRO ’52. ACM, 2019.

\bibitem[\protect\citeauthoryear{Zhuang \bgroup \em et al.\egroup }{2023}]{zhuang2023}
Weiming Zhuang, Chen Chen, and Lingjuan Lyu.
\newblock When foundation model meets federated learning: Motivations, challenges, and future directions, 2023.

\bibitem[\protect\citeauthoryear{Ziller \bgroup \em et al.\egroup }{2021}]{Ziller2021_pysyft}
Alexander Ziller, Andrew Trask, et~al.
\newblock {\em PySyft: A Library for Easy Federated Learning}.
\newblock Springer International Publishing, 2021.

\end{thebibliography}
